\documentclass[a4paper]{article}

\usepackage{INTERSPEECH2019}
\usepackage{amssymb,amsmath,graphicx,epsfig,color,cite,multirow}
\usepackage{standalone}
\usepackage{siunitx}
\usepackage[acronym]{glossaries}
\usepackage{balance}
\usepackage{url}

\usepackage[english,capitalize]{cleveref}
\usepackage{tikz}
\usetikzlibrary{positioning,chains,calc,shapes.geometric, shadows, shapes.misc}

\newacronym{SE}{SE}{speech enhancement}
\newacronym{WPE}{WPE}{Weighted Prediction Error}
\newacronym{NN}{NN}{Neural Network}
\newacronym{SAD}{SAD}{Speech Activity Detection}
\newacronym{GSS}{GSS}{Guided Source Separation}
\newacronym{BAN}{BAN}{Blind Analytic Normalization}

\title{Guided Source Separation Meets a Strong ASR Backend:\\Hitachi/Paderborn University Joint Investigation for Dinner Party ASR}
\name{Naoyuki Kanda$^{1,*}$, Christoph Boeddeker$^{2,*}$, Jens Heitkaemper$^{2,*}$, \\
Yusuke Fujita$^1$, Shota Horiguchi$^1$, Kenji Nagamatsu$^1$, Reinhold Haeb-Umbach$^2$}
\address{
  $^1$Hitachi Ltd., Japan\\
  $^2$Paderborn University, Germany}
\email{naoyuki.kanda.kn@hitachi.com, boeddeker@nt.upb.de, heitkaemper@nt.upb.de}

\begin{document}
\maketitle

\begin{abstract}
In this paper, we present Hitachi and Paderborn University's joint effort for automatic speech recognition (ASR) in a dinner party scenario. 
The main challenges of ASR systems for dinner party recordings obtained by multiple microphone arrays are (1)  heavy speech overlaps, 
(2) severe noise and reverberation, (3) very natural conversational content, 
and possibly (4) insufficient training data. 
As an example of a dinner party scenario, we have chosen the data presented during the CHiME-5 speech recognition challenge, 
where the baseline ASR had a 73.3\% word error rate (WER), 
and even the best performing system at the CHiME-5 challenge had a 46.1\% WER. 
We extensively investigated a combination of the guided source separation-based speech enhancement technique 
and an already proposed strong ASR backend and found that a tight combination of these techniques provided substantial accuracy improvements.
 Our final system achieved WERs of \textcolor{black}{39.94\% and 41.64\%} for the development and evaluation data, respectively, 
both of which are the best published results for the dataset. 
We also investigated with additional training data on the official small data in the CHiME-5 corpus 
to assess the intrinsic difficulty of this ASR task.
\end{abstract}

\section{Introduction}
\renewcommand{\thefootnote}{\fnsymbol{footnote}}
\footnotetext{* Equal contribution}
\renewcommand{\thefootnote}{\arabic{footnote}}
Due to recent advances in deep learning \cite{seide2011conversational,dahl2012context,hinton2012deep}, 
the word error rates (WERs) of automatic speech recognition (ASR) for some datasets have become close to 
(Switchboard \cite{xiong2016achieving,saon2017english}) or just below (LibriSpeech \cite{amodei2016deep} and \cite{kanda-interspeech2018}) 
the WER level of human transcribers. 
However, despite this progress, noise and reverberation still severely increase the WERs. 
In particular, multi-talker speech recognition is one of the most difficult settings 
for speech recognition \cite{yoshioka2018recognizing,barker2018fifth,kanda2018hitachi} 
because of the difficulty of separating the target speech signal from other interfering speech signals. 
One example is meeting speech recognition, where it is known that the WERs are 
still around 30\% \cite{yoshioka2018recognizing,kanda2019icassp} even with state-of-the-art speech recognizers. 
Another example is distant speech recognition in a daily home environment, 
such as a dinner party \cite{barker2018fifth}, which will be useful for developing intelligent home devices.

To push the boundary of the current state-of-the-art ASR for such difficult noisy environments, 
the CHiME challenge has been held every one or two years \cite{barker2013pascal,vincent2013second,barker2015third,barker2018fifth}. 
In the latest CHiME-5 challenge \cite{barker2018fifth}, dinner party recordings with four participants were provided. 
The recordings were conducted with six microphone arrays, each of which had four microphones. 
The ASR for this dataset was significantly more difficult 
compared with the previous challenge \cite{barker2013pascal,vincent2013second,barker2015third} 
because of (1)  heavy speech overlaps, (2) severe noise and reverberation, (3) very natural conversational content, 
and possibly (4) insufficient training data. 
The first to third reasons came from the nature of the recordings. 
On the other hand, the fourth reason came from the regulation of the challenge 
in which only 40 hours of official training data was allowed to be used 
for the official challenge system\footnote{32 microphones were used for the recordings, 
so the total duration of the data was about 1,300 hours.}. 
As a result, the baseline system had a 73.3\% WER \cite{barker2018fifth}, 
and even the best performing system \cite{du2018ustc} achieved a 46.1\% WER.

At the time of the challenge, Hitachi provided many contributions with Johns Hopkins University (JHU) on acoustic modeling (AM), 
language modeling (LM), and decoding techniques and achieved the second best result of a 48.2\% WER \cite{kanda2018hitachi}. 
On the other hand, Paderborn University achieved very promising speech enhancement (SE) techniques, 
named guided source separation (GSS)\footnote{\url{https://github.com/fgnt/pb_chime5}}, 
which 
achieved a significant improvement 
for evaluation data in multiple array settings \cite{boeddecker2018chime5,kitza2018rwth}. 
We thought this is worth investigating to evaluate the results combining our contributions 
to assess the state-of-the-art performance of today's ASR system. 

According to the discussion above, in this paper, we present Hitachi and Paderborn University's joint effort 
on developing a state-of-the-art ASR system on the CHiME-5 corpus. 
We conducted investigations from two perspectives. 
Firstly, we conducted a comprehensive investigation of the system that utilizes all the contributions 
we separately proposed in the CHiME-5 challenge. 
By tightly combining our contributions, we achieved the new best records for the dataset. 
Secondly, we addressed the concern about the data scarcity problem by using AM or LM with more training data. 
We believe our results will provide better insights into the intrinsic difficulty of this ASR task. 

\section{CHiME-5 Corpus}
\label{sec:corpus}
The CHiME-5 database contains recordings of dinner parties attended by four friends 
who \textcolor{black}{engaged} in casual conversations. 
Each party \textcolor{black}{was} split into three parts\textcolor{black}{:} preparing food, dining, and socializing. 
All the parts \textcolor{black}{took} place in different rooms and \textcolor{black}{lasted} at least 30 minutes. 
\textcolor{black}{Recordings were conducted with} six Microsoft Kinect\textregistered{} microphone arrays 
with four audio channels each and two arrays per room. 
For all the parties, every speaker wore two in-ear microphones. 
These in-ear microphone signals were considered as close talk, 
and they were only used in training and development.

The training dataset comprises about 40 hours of audio, while the development and evaluation set consist of five hours each. 
Because of the natural conversation style, around $\SI{22}{\%}$ of the signal recorded for the training set 
includes more than one active speaker. 
For the development and evaluation set\textcolor{black}{,} this number is around $\SI{40}{\%}$ and $\SI{25}{\%}$, respectively. 
Due to the great difficulty of the ASR task, annotations regarding the start and end times for each utterance 
were allowed to be used at the time of the CHiME-5 challenge, and we also utilized the same time annotations in this study.

\section{Speech enhancement}
\label{sec:enhancement}
In this study, we applied the \gls{SE} proposed by the Paderborn University team for the CHiME-5 challenge \cite{boeddecker2018chime5}. 
The system uses spatial mixture models, which are learned in an unsupervised fashion. 
The time annotations 
in the database 
are algorithmically fine-tuned to obtain source activity information at word level precision. 
This 
time annotation
is used to guide the source separation process. 
\begin{figure}[t]
  \centering
  \tikzset{%
  block/.style    = {draw, thick, rectangle, minimum height = 1.5em, minimum width = 3em, rounded corners=0.3em, fill=black!6},
  sum/.style      = {draw, circle, node distance = 2cm}, 
  cross/.style={path picture={\draw[black](path picture bounding box.south east) -- (path picture bounding box.north west)
		 (path picture bounding box.south west) -- (path picture bounding box.north east);}}
               }
\tikzstyle{branch}=[{circle,inner sep=0pt,minimum size=0.3em,fill=black}]

\begin{tikzpicture}[auto, line width=0.1em, node distance = 1cm]

\tikzset{pics/.cd,
	pic switch/.style args={#1 times #2}{code={
			\tikzset{x=#1/2,y=#2/2}
			\coordinate (-north west) at (-1,1);
			\coordinate (-north east) at (1,1);
			\coordinate (-south west) at (-1,-1);
			\coordinate (-south east) at (1,-1);
			\coordinate (-north) at (0,1);
			\coordinate (-east) at (1,0);
			\coordinate (-south) at (0,-1);
			\coordinate (-west) at (-1,0);
			
			\draw [line cap=rect] (-1,1) -- (-1,0.5);
			\draw [line cap=rect] (-1,-1) -- (-1,-0.5);
			\draw [line cap=round] (1, 0) -- ($(1, 0)!2/3!(-1.3,-0.8)$);
			\draw [line cap=rect] ($(1, 0)!1/3!(-1.3,-0.8)$) -- (-1.3,-0.8);
	}}
}

\tikzset{>=stealth}
\tikzstyle{arrow}=[{}-{>}]

\node[block, align=center, at={(2.2,0)}](WPE){WPE};
\node[block, align=center, anchor=south west](BSS) at ($(WPE.north east) + (0.2, 0.3)$) {GSS};
\node[block, align=center, anchor=west](BF) at ($(BSS.south east |- WPE.east) + (0.2, 0)$) {BF};
\node[block, align=center, anchor=west] at ($(BF.center) + (1.2, 0)$) (ASR){ASR};

\pic [] (switch) at ($(BSS.west) + (-0.5, 0)$) {pic switch={1.5em times 1.5em}};


\coordinate(Y) at (0, 0);
\coordinate(a) at ($(switch-south west -| Y)$);

\draw[arrow] (Y) node[above right, align=left]{STFT} -- node[strike out,draw, anchor=center,pos=0.8,-]{} node[above=0.06,pos=0.8]{\small{$N$}}(WPE);
\draw[arrow] (WPE) -| node[strike out,draw, pos=0.3,-, anchor=center]{} node[above=0.06, pos=0.3]{\small{$N$}} (WPE -| BSS) node[branch]{} -- (BSS);
\draw[arrow] (WPE) -- (BF);
\draw[] (a) node[above right,align=left]{Time\\annotation} -- (switch-south west);
\draw[arrow] (BSS.east) node[above right]{Mask} -| (BF);
\draw[arrow] (BF) -- (ASR);


\draw[] (ASR.east) -++ (0.5, 0) -- ($(ASR.east |- BSS.north) + (0.5, 0.2)$) node[below left]{Alignment} -| ($(switch-north west) + (-0.2, 0)$) -- (switch-north west);

\draw[arrow] (switch-east) -- (BSS);


\node[right] at (switch-south west){\footnotesize{init}};


\end{tikzpicture}
  \vspace{-3mm}
  \caption{Overview of speech enhancement system}
  \label{fig:enhancement_block}
\vspace{-5mm}
\end{figure}
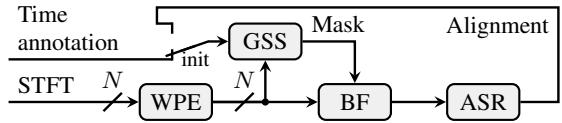

An overview of this system is shown in \cref{fig:enhancement_block}. 
The \gls{SE} combines \gls{WPE}~\cite{yoshioka2012GWPE,drude2018naraWPE} for dereverberation 
with statistical beamforming (BF) for source extraction (MVDR beamformer~\cite{souden2010MVDR,erdogan2016MVDR} 
with a \glsdesc{BAN} postfilter~\cite{warsitz2007GEVBAN}). 
The target and distortion masks for the beamformer are estimated from a \gls{GSS} system 
consisting of a spatial mixture model using complex angular central Gaussian distributions~\cite{ito2016cACGMM}. 
The \gls{GSS} makes efficient use of the utterance start and end time annotations found in the database as follows. 
First, \textcolor{black}{GSS} determines the number of active speakers from \textcolor{black}{the time annotations}. 
Second, \textcolor{black}{GSS} uses \textcolor{black}{the time annotations} to initialize the posterior probability 
of a source being active as one divided by the number of active speakers in a time-frame. 
Third, the posterior probability of a speaker is fixed to be zero 
whenever the speaker is inactive according to the \textcolor{black}{time} annotations.

Using the \textcolor{black}{time annotation} in the described fashion 
eliminates the need to estimate the number of active speakers. 
Furthermore, the iterative estimation of the posterior probabilities encourages a permutation-free solution 
and is guided to keep it free of permutations because if the source activity pattern between all active sources is sufficiently different, 
the posterior will tend to be permutation free. 
This includes the absence of permutations between frequencies and between utterances; 
furthermore, it avoids the situation where a source is modeled by more than one mixture component. 
Since the activity patterns of the sources in an utterance may not always be sufficiently different 
\textcolor{black}{(}e.g., if two speakers are simultaneously active during the whole utterance\textcolor{black}{)}, 
the utterance is extended with a \SI{15}{\second} left and right context. 
An in depth description of the \gls{SE} system can be found in Boeddeker et al.'s study \cite{boeddecker2018chime5}.

However, the annotations provided by the database are not perfect. 
For example, the silence frames at the start and end of an utterance or between words are not marked. 
We therefore \textcolor{black}{fine-tuned the annotations}. 
In \textcolor{black}{the previous study \cite{boeddecker2018chime5},} a source activity detector (SAD) \textcolor{black}{neural network} 
was trained \textcolor{black}{and used} to predict the activity of a source from the observations, the time annotations, and a mask from GSS.
 \textcolor{black}{The training data for SAD was} obtained from the forced alignment on the in-ear microphone signals computed by ASR. 
\textcolor{black}{On the other hand,} \textcolor{black}{a strong} ASR system can itself estimate a good alignment on the enhanced test data. 
The procedure \textcolor{black}{we finally applied in this study} is as follows. 
First, the data is enhanced \textcolor{black}{by using SE with the SAD-based annotation}. 
Next, the ASR estimates the alignment on the enhanced signals. 
Then, the time annotations of the database are adjusted to be zero where the alignments indicate silence. 
With this refined guiding information, the enhancement is repeated, followed by the final recognition pass.

Note the whole enhancement system is independent of the number of input channels $N$. 
It can be applied on the reference array ($N=4$) or jointly on all arrays ($N=24$). 
\textcolor{black}{In theory, stacking all arrays into one big array could improve the performance (e.g., being more spatially discriminable) 
or could degrade the performance (e.g., the arrays are not perfectly synchronized).}
 \textcolor{black}{However, in} the experiments, we saw a large benefit from \textcolor{black}{stacking} all array data.

\section{Acoustic modeling}
\label{sec:am}

\begin{table*}[t]
\centering
\caption{WERs (\%) for development and evaluation sets with various settings.}
\vspace{-3mm}
\label{tab:compare_techniques}
{\scriptsize
\begin{tabular}{lclcccccc}
\toprule
System & Array   &   \multicolumn{1}{c}{SE} &Array Combination      & AM   &RNN-LM & HD & DEV (\%) & EVAL (\%) \\ 
\cmidrule{1-9}
Baseline  \cite{barker2018fifth} & Single   & BeamFormIt \cite{anguera2007acoustic} & -& 1-AM     &  &  & 81.1 & 73.3 \\ 
USTC/iFlytek \cite{du2018ustc} & Single   & Multi-stage BF \cite{du2018ustc}&    -                  & 5-AM & \checkmark &  & 50.2 & 46.1\\ 
USTC/iFlytek \cite{du2018ustc} & Multiple   & Multi-stage BF \cite{du2018ustc} &  Array Selection \cite{du2018ustc} & 5-AM & \checkmark &  & 45.0 & 46.1\\ 
\cmidrule{1-9}
Our System 1 & Single   & RAW & -                     & 1-AM &  &  & 63.45 & -\\
Our System 2 & Single   & WPE & -                     & 1-AM &  &  & 63.05 & -\\
Our System 3 & Single   & WPE + CGMM-MVDR\cite{kanda2018hitachi} & -    & 1-AM &  &  & 62.24 & -\\
Our System 4 & Single   & WPE + SA-NN-MVDR\cite{kanda2018hitachi} & -   & 1-AM &  &  & 61.91 & -\\
Our System 5 & Single   & WPE + GSS + BF w/ Context\cite{boeddecker2018chime5}   & -          & 1-AM &  &  & 58.57 & -\\ 
Our System 6 & Single   & WPE + GSS w/ SAD + BF w/ Context\cite{boeddecker2018chime5} & -          & 1-AM &  &  & 58.05 &-\\ 
Our System 7 & Single   & WPE + GSS w/ SAD + BF w/o Context & -          & 1-AM &  &  & 58.13  & 53.76 \\ 
Our System 8 & Single   & WPE + GSS w/ ASR + BF w/o Context & -          & 1-AM &  &  & 58.29  & 53.10 \\ 
Our System 9 & Single   & WPE + GSS w/ ASR + BF w/o Context& -          & 6-AM &  &  & 54.62 & 49.18\\ 
Our System 10 & Single   & WPE + GSS w/ ASR + BF w/o Context& -          & 6-AM & \checkmark &   & 53.18 & 47.54\\
Our System 11 & Single   & WPE + GSS w/ ASR + BF w/o Context& -          & 6-AM & \checkmark & \checkmark & 52.07  & 47.31 \\ 
\cmidrule{1-9}
Our System 12 & Multiple & WPE + SA-NN-MVDR\cite{kanda2018hitachi} & ROVER  & 1-AM  &  &  & 57.50 &-\\
Our System 13 & Multiple & WPE + GSS + BF w/ Context\cite{boeddecker2018chime5}  & Stacking in SE & 1-AM &   &  & 50.23 &-\\ 
Our System 14 & Multiple & WPE + GSS w/ SAD + BF w/ Context\cite{boeddecker2018chime5} & Stacking in SE & 1-AM &   &  & 49.21 &- \\ 
Our System 15 & Multiple & WPE + GSS w/ SAD + BF w/o Context & Stacking in SE & 1-AM &   &  & 46.54 & 51.99\\ 
Our System 16 & Multiple & WPE + GSS w/ ASR + BF w/o Context & Stacking in SE & 1-AM &   &  & 45.14 & 47.29 \\ 
Our System 17 & Multiple & WPE + GSS w/ ASR + BF w/o Context & Stacking in SE & 6-AM &   &  & 41.67 & 43.70 \\
Our System 18 & Multiple & WPE + GSS w/ ASR + BF w/o Context & Stacking in SE & 6-AM & \checkmark &  & {\bf 39.94} & {\bf 41.64}\\
Our System 19 & Multiple & WPE + GSS w/ ASR + BF w/o Context & Stacking in SE & 6-AM & \checkmark & \checkmark & 40.26 & 42.00\\
\bottomrule
\end{tabular}
}
\vspace{-5mm}
\end{table*}

\begin{figure}[t]
\centerline{\epsfig{figure=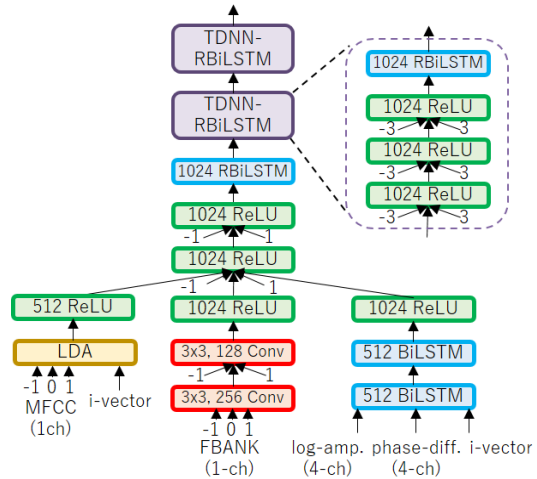,width=70mm}}
\vspace{-3mm}
\caption{Overview of CNN-TDNN-RBiLSTM acoustic model}
\label{fig:am}
\vspace{-5mm}
\end{figure}

In this study, we applied the AM with single- and multi-channel input branches 
proposed for the Hitachi/JHU system \cite{kanda2018hitachi} that provided us with substantial improvement over the baseline AM. 
An overview of the acoustic model is depicted in \cref{fig:am}. 
Our AM consists of a convolutional neural network (CNN), time delay neural network (TDNN), 
and our proposed residual bidirectional long short-term memory (RBiLSTM) \cite{kanda2018hitachi}. 
The unique part of this model architecture is in its input branch. 
This model has input branches for single-channel features and an input branch that accepts multi-channel features. 
The multi-channel input branch acts as a learnable SE module. 
On the other hand, the single-channel input branch is used to accept enhanced speech by using a complementary SE module. 
By having these two types of input branches, our model uniquely has the ability to use a complementary SE module 
while exploiting the power of jointly trained AM and SE architecture. 

We use mel-frequency cepstral coefficients (MFCCs) and log mel-filterbank (FBANK) as input for the single-channel branch. 
On the other hand, we use two types of features that represent multi-channel input signals for the multi-channel branch. 
One type of feature is the log amplitude for each microphone, and the other type of feature is the phase difference 
between each microphone and the first microphone. 
We trained the AM by using LF-MMI criterion \cite{povey2016purely} and then further updated the AM 
by using lattice-free state-level minimum Bayes risk (LF-sMBR) criterion \cite{kanda-interspeech2018}. 
The details of our training schemes and comprehensive investigation results can be found 
in previous studies \cite{kanda2018hitachi,kanda2019icassp}.

\section{Language modeling and decoding}
\label{sec:lm}
In this study, we basically followed the language modeling and decoding procedure 
proposed for the Hitachi/JHU system \cite{kanda2018hitachi}. 
We trained recurrent neural network language models (RNN-LMs) 
by using the official transcription of the training data. 
We prepared two 2-layer LSTM-based models with forward and backward direction. 
The average score of the official n-gram LM, forward RNN-LM, and backward RNN-LM was used with a weighting of 0.5:0.25:0.25. 

In the decoding phase, we used the N-best ROVER method \cite{fiscus1997post} to combine the results from different AMs. 
For the AM combination, we trained six types of AMs: \{CNN-TDNN-RBiLSTM, CNN-TDNN-LSTM, CNN-TDNN-BiLSTM\} x \{3500, 7000\} senones. 
In the Hitachi/JHU system \cite{kanda2018hitachi}, the recognition results from different microphone arrays were also combined by ROVER. 
However, this technique was only effective for the development set, and no gain was observed for the evaluation set \cite{kanda2018hitachi}. 
Instead, in this study, we exploited information from multiple arrays at the stage of SE, as described in Section \ref{sec:enhancement}. 
Therefore, we omitted the ROVER-based array combination when we used the GSS technique. 

We also applied the ``hypothesis deduplication (HD)'' proposed for the Hitachi/JHU system \cite{kanda2018hitachi}. 
In HD, if the same words were recognized for overlapping utterances, 
words with lower confidence were excluded from the hypothesis.

\section{Evaluation}
\label{sec:eval}
\subsection{Experimental settings}
In our evaluation, we used the CHiME-5 corpus, the overview of which is described in Section \ref{sec:corpus}. 
Unless otherwise specified, we followed the regulations in the CHiME-5 challenge 
where only the official training data was allowed for 
AM and LM
training. 
The original duration of the 
training data was 40.6 hours. 
When we trained our AMs, we applied speed and volume perturbation \cite{ko2015audio}, 
reverberation and noise perturbation \cite{ko2017study}, and bandpass perturbation \cite{kanda2018hitachi}, 
which produced roughly 4,500 hours of training data. 
Further details of the training pipeline for our AM are described in our previous study \cite{kanda2018hitachi}. 

The duration of development data (DEV) and evaluation data (EVAL) were 4.5 hours and 5.2 hours, respectively. 
There were two official tasks for the dataset; one used only reference array data (single array track), 
and the other one used all the arrays (multiple array track). 
For both tasks, all the parameters were tuned by development set, and the best parameters were used for decoding the evaluation set.

\subsection{Results of Hitachi/Paderborn University joint system}
The results of our joint system are presented in \cref{tab:compare_techniques}. 
The first row shows the result 
of
the CHiME-5 baseline system \cite{barker2018fifth}, 
and the second and third rows show the results 
of
the best system at the CHiME-5 challenge \cite{du2018ustc}.

We firstly evaluated our system in the single-array setting. 
System 1 to system 4 are the systems without or with the SE techniques proposed for the Hitachi/JHU system \cite{kanda2018hitachi}. 
Then, by applying GSS, we achieved a 3.34\% WER reduction (system 5). 
The addition of SAD further improved the accuracy, and we achieved a 58.05\% WER (system 6). 
We also tried to remove context information in beamforming (system 7) 
and use the alignment information produced by system 7 for replacing SAD (system 8). 
Although the last two changes had almost no impact on the single-array setting, 
they significantly improved the WER for the multiple-array setting (discussed in the next paragraph), 
so we selected the SE settings of system 8 for the final system for consistency. 
Finally, by applying various decoding techniques, such as AM combination (system 9), RNN-LM (system 10), and HD (system 11), 
the WER was significantly improved, and we achieved \textcolor{black}{52.07\% and 47.31\% WERs for DEV and EVAL, respectively}. 

Next, we evaluated our joint system in the multiple-array settings. 
Firstly, we show the result with the SE techniques proposed for the Hitachi/JHU system \cite{kanda2018hitachi} (system 12). 
GSS again significantly improved the accuracy and achieved a 50.23\% WER (system 13). 
By adding SAD, the WER was further improved to 49.21\% (system 14). 
We found that removing context information in beamforming significantly improved the accuracy (system 15). 
\textcolor{black}{This gain may be traced back to an improved statistical estimation if just the utterance is considered, 
which allows us to ignore cross talkers that are only active during the context 
and make a better modeling of moving speakers.} 
In addition, we replaced SAD information by using the alignment produced by system 15, 
which gave us a significant WER improvement (system 16). 
Finally, by applying the decoding techniques of AM combination (system 17) and RNN-LM (system 18), 
we achieved the best result of \textcolor{black}{39.94}\% and 41.64\% WERs for DEV and EVAL, respectively. 
Interestingly, HD degraded the WER for the multiple array settings (system 19). 
This implies that most of the cross talks were removed by GSS.

\textcolor{black}{One notable result for this experiment is the improvement by using multiple arrays when we used GSS for speech enhancement.
 The WER was improved from 58.13\% to 46.54\% (11.59\% absolute improvement) when we used 6 arrays for GSS (system 7 and 15) 
while the WER was improved by only 4.41\% when we combined the results from each array by using ROVER (system 4 and 12) 
as the Hitachi/JHU system did \cite{kanda2018hitachi}. 
To assess the effect of the number of arrays, we conducted the experiment with various numbers of arrays, 
the results of which are shown in \cref{tab:compare_array_num}. 
We found that the WER improvement was not saturated even when we used 6 arrays, 
and we can expect further WER improvement with a larger number of microphone arrays. 
We also found that it is better to remove context information in the BF calculation when we use multiple arrays.} 

\begin{table}[t]
\centering
\caption{WERs (\%) for development set with different numbers of arrays for GSS. The CNN-TDNN-RBiLSTM-AM and the official LM were used for decoding.}
\vspace{-3mm}
\label{tab:compare_array_num}
{\footnotesize
\begin{tabular}{ccc}
\toprule
Arrays & \multicolumn{2}{c}{Context in BF} \\ 
     & On & Off\\
\cmidrule{1-3}
1 & 58.05 & 58.13 \\
3 & 52.30 & 48.81 \\
6 & 49.21 & 46.54 \\
\bottomrule
\end{tabular}
}
\vspace{-3mm}
\end{table}

\begin{table}[t]
\centering
\caption{WERs (\%) for our best system (\#11 and \#18 in \cref{tab:compare_techniques}).}
\vspace{-3mm}
\label{tab:best_system}
{\scriptsize
\begin{tabular}{c|c|c|c|c|c|c}
\toprule
Track & \multicolumn{2}{|c|}{Session}  & Kitchen & Dining & Living & Overall\\
\midrule
\multirow{4}{*}{\footnotesize Single}
& \multirow{2}{*}{Dev}  & S02 & 62.33 & 52.82 & 44.62 & \multirow{2}{*}{52.07} \\
&                       & S09 & 51.87 & 54.02 & 48.09 & \\ \cmidrule{2-7}
& \multirow{2}{*}{Eval} & S01 & 60.07 & 40.88 & 60.94 & \multirow{2}{*}{47.31}\\
&                       & S21 & 49.09 & 38.14 & 42.67 & \\
 \midrule
\multirow{4}{*}{\footnotesize Multiple}
& \multirow{2}{*}{Dev}  & S02 & 46.66 & 45.07 & 36.19 & \multirow{2}{*}{39.94} \\
&                       & S09 & 36.40 & 39.43 & 35.33 & \\ \cmidrule{2-7}
& \multirow{2}{*}{Eval} & S01 & 53.93 & 35.66 & 49.78 & \multirow{2}{*}{41.64} \\
&                       & S21 & 46.43 & 34.53 & 36.64 & \\
\bottomrule
\end{tabular}
}
\vspace{-5mm}
\end{table}

In \cref{tab:best_system}, we show the detailed results of our best system for single array and multiple array track. 
To the best of our knowledge, our multiple-array results are 
the best 
published results for this dataset.

\subsection{Investigation with larger training data}
So far, we followed the regulations of CHiME-5. 
However, the original duration of the training data was only about 40 hours, 
and it is unclear whether the difficulty of this ASR task came from insufficient training data 
or the intrinsic property of this dataset. 
Therefore, in this section, we conducted the evaluation with larger training data.

\subsubsection{Comparison with AM using larger dataset}
We firstly conducted the evaluation with a very strong AM, 
which once achieved the best published results \cite{kanda-interspeech2018} for LibriSpeech corpus \cite{panayotov2015librispeech}. 
The training data was originally 960 hours of LibriSpeech corpus, 
and 
it 
was further augmented to roughly 3,000 hours by volume
and speed perturbation. 
The AM consists of CNN, LSTM, and TDNN and was trained by LF-sMBR. 
Please refer to 
the paper
\cite{kanda-interspeech2018} for further details. 

The result for this LibriSpeech AM was shown in the first line of \cref{tab:compare_am}. 
In the second and third lines, the results of the baseline AM 
and our best AM (CNN-TDNN-RBiLSM) 
were shown, respectively. 
As shown in the table, LibriSpeech AM produced the worst WER of 62.09\% even with 960 hours of training data and a very strong SE module. 
Of course, there could be a better way of using the large data; e.g. we could use the data for pretraining. 
Nonetheless, we can at least say that this ASR task is very difficult regardless of the data size for AMs, 
and the naive use of 960 hours of training data was much worse than using the matched 40 hours of training data. 

\begin{table}[t]
\centering
\caption{WERs (\%) for development set with different AMs. The multiple-array GSS was used with official LM.}
\vspace{-3mm}
\label{tab:compare_am}
{\footnotesize
\begin{tabular}{ccc}
\toprule
 AM & Training Data & DEV (\%) \\
\cmidrule{1-3}
CNN-TDNN-LSTM \cite{kanda-interspeech2018} & LibriSpeech (960h)  & 62.09 \\
Baseline TDNN & CHiME-5 (40h) & 58.39 \\
CNN-TDNN-RBiLSTM & CHiME-5 (40h) & 45.14 \\ 
\bottomrule
\end{tabular}
}
\vspace{-3mm}
\end{table}

\subsubsection{Comparison with LMs using larger dataset}
Finally, we compared various LMs with larger training data. 
In this experiment, we used the transcriptions in the AMI meeting corpus \cite{carletta2005ami} 
and LibriSpeech corpus \cite{panayotov2015librispeech} for the training data. 
The number of words in the transcriptions of the CHiME-5, AMI, and LibriSpeech corpus were 0.44M, 0.80M, and 9.40M, respectively. 
We trained 3-gram LMs with Kneser-Ney smoothing \cite{kneser1995improved} and interpolated them 
with the 3-gram LM trained with CHiME-5 transcription. 
For the model interpolation, we used MIT-LM \footnote{\url{https://github.com/mitlm/mitlm}}, 
and the interpolation weights were tuned by using the transcription of the CHiME-5 development data.

The perplexity (PPL) and WER are listed in \cref{tab:compare_lm}. 
In the case of LM, the larger the data, the better the results, and the best LM achieved a 24 point better PPL of 131. 
However, the WER improvement obtained by using this LM was only \textcolor{black}{0.93}\%. 
According to these results, we concluded that the difficulty of this ASR task 
mainly came from its intrinsic property rather than insufficient training data. 

\begin{table}[t]
\centering
\caption{Comparison of LMs. The multiple-array-based GSS and CNN-TDNN-RBiLSTM AM was used for decoding.}
\vspace{-3mm}
\label{tab:compare_lm}
{\footnotesize
\begin{tabular}{lc|cc}
\toprule
Training Data  & \# of Words  & \multicolumn{2}{c}{DEV} \\ 
   & &  PPL & WER (\%)  \\ 
\cmidrule{1-4}
C (Baseline)     & 0.4M    & 155 & 45.14   \\
C + A  & 1.2M   & 140 & 45.10 \\
C + L  & 9.8M  & 134 & 44.49 \\
C + A + L & 10.6M & 131 & 44.21 \\
\bottomrule
\end{tabular}
}
\\\footnotesize{C: CHiME-5, A: AMI, L:LibriSpeech}
\vspace{-5mm}
\end{table}

\section{Conclusion}
\label{sec:conclusion}
In this paper, we presented Hitachi and Paderborn University's joint effort on ASR for the CHiME-5 speech corpus. 
We gathered our contributions, which were separately proposed at the CHiME-5 challenge, 
and our best system finally achieved WERs of 39.94\% and 41.64\% for development and evaluation data, respectively, 
both of which are the best records for the dataset. 
We also conducted investigations with larger training data for AM and LM. 
We found that simply using larger data had no impact or a marginal impact on the WER, 
which indicated the intrinsic difficulty of this ASR task.

\section{Acknowledgements}
We deeply thank Prof. Shinji Watanabe for connecting Hitachi and Paderborn University for this great collaboration.

This work was in part supported by DFG under contract number Ha3455/14-1. The computational resources were provided by the Paderborn Center for Parallel Computing.

\balance
\bibliographystyle{IEEEtran}
\bibliography{thesis}

\end{document}